\documentclass[10pt]{article} %
\usepackage[accepted]{rlc}
\usepackage{shortcuts}

\usepackage{amssymb}            %
\usepackage{mathtools}          %
\usepackage{mathrsfs}           %
\mathtoolsset{showonlyrefs}     %
\usepackage{graphicx}           %
\usepackage{subcaption}         %
\usepackage[space]{grffile}     %
\usepackage{url}                %
\usepackage{algorithm}
\usepackage{algpseudocode}
\usepackage{wrapfig}
\usepackage{booktabs}

\title{RL for Consistency Models: Faster Reward Guided Text-to-Image Generation}

\author{%
  Owen Oertell \\
  Department of Computer Science\\
  Cornell University\\
  \texttt{ojo2@cornell.edu} \\
  \And
  Jonathan D. Chang \\
  Department of Computer Science\\
  Cornell University\\
  \texttt{jdc396@cornell.edu} \\
  \And
  Yiyi Zhang \\
  Department of Computer Science\\
  Cornell University\\
  \texttt{yz2364@cornell.edu} \\
  \And
  Kianté Brantley \\
  Department of Computer Science\\
  Cornell University\\
  \texttt{kdb82@cornell.edu} \\
  \And
  Wen Sun \\
  Department of Computer Science\\
  Cornell University\\
  \texttt{ws455@cornell.edu} \\
}

\definecolor{dark-green}{rgb}{0,0.7,0}

\newcommand{\alg}{RLCM}
\newcommand{\ddpo}{DDPO}

\begin{document}

\maketitle

\begin{abstract}
Reinforcement learning (RL) has improved guided image generation with diffusion models by directly optimizing rewards that capture image quality, aesthetics, and instruction following capabilities. However, the resulting generative policies inherit the same iterative sampling process of diffusion models that causes slow generation. To overcome this limitation, consistency models proposed learning a new class of generative models that directly map noise to data, resulting in a model that can generate an image in as few as one sampling iteration. In this work, to optimize text-to-image generative models for task specific rewards and enable fast training and inference, we propose a framework for fine-tuning consistency models via RL. Our framework, called Reinforcement Learning for Consistency Model (\alg), frames the iterative inference process of a consistency model as an RL procedure. 
Comparing to RL finetuned diffusion models, \alg{} trains significantly faster, improves the quality of the generation measured under the reward objectives, and speeds up the inference procedure by generating high quality images with as few as two inference steps.
Experimentally, we show that \alg{} can adapt text-to-image consistency models to objectives that are challenging to express with prompting, such as image compressibility, and those derived from human feedback, such as aesthetic quality. Our code is available at \url{https://rlcm.owenoertell.com}.
\end{abstract}

\begin{figure}[t]
    \centering
    \includegraphics[width=\textwidth]{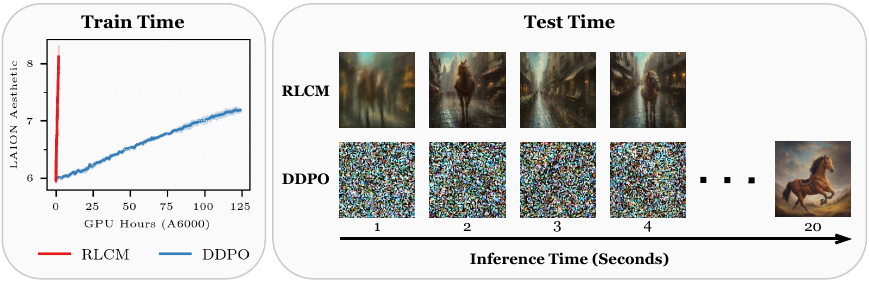}
    \caption{\textbf{Reinforcement Learning for Consistency Models (\alg{})}. We propose a new framework for finetuning consistency models using RL. On the task of optimizing aesthetic scores of a generated image, comparing to a baseline which uses RL to fine-tune diffusion models (DDPO), \alg{} trains (left) and generates images (right) significantly faster, with higher image quality measured under the aesthetic score. Images generated with a batch size of 8 and \alg{} horizon set to 8.} 
    \label{fig:front-page-figure}
    \vspace{-5px}
\end{figure}

\section{Introduction}
\label{sec:introduction}

Diffusion models have gained widespread recognition for their high performance in various tasks, including drug design \citep{xu2022geodiff} and control \citep{janner2022planning}. In the text-to-image generation community, diffusion models have gained significant popularity due to their ability to output realistic images via prompting. Despite their success, diffusion models in text-to-image tasks face two key challenges. First, generating the desired images can be difficult for downstream tasks whose goals are hard to specify via prompting. Second, the slow inference speed of diffusion models poses a barrier, making the iterative process of prompt tuning computationally intensive.

To enhance the generation alignment with specific prompts, diffusion model inference can be framed as sequential decision-making processes, permitting the application of reinforcement learning (RL) methods to image generation~\citep{black2024ddpo, fan2023DPOK}. The objective of RL-based diffusion training is to fine-tune a diffusion model to maximize a reward function directly that corresponds to the desirable property.

Diffusion models also suffer from slow inference since they must take many steps to produce competitive results. This leads to slow inference time and even slower training time. Even further, as a result of the number of steps we must take, the resultant Markov decision process (MDP) possesses a long time horizon which can be hard for RL algorithms optimize. To resolve this, we look to consistency models. These models directly map noise to data and typically require only a few steps to produce good looking results. Although these models can be used for single step inference, to generate high quality samples, there exits a few step iterative inference process which we focus on. Framing consistency model inference, instead of diffusion model inference, as an MDP admits a much shorter horizon. This enables faster RL training and allows for generating high quality images with just few step inference.

More formally, we propose a framework \textbf{R}einforcement \textbf{L}earning for \textbf{C}onsistency \textbf{M}odels (\alg{}), a framework that models the inference procedure of a consistency model as a multi-step Markov Decision Process,  allowing one to fine-tune consistency models toward a downstream task using just a reward function. Algorithmically, we instantiate RLCM using a policy gradient algorithm as this allows for optimizing general reward functions that may not be differentiable. %
In experiments, we compare to the current more general method, DDPO \citep{black2024ddpo} which uses policy gradient methods to optimize a diffusion model. In particular, we show that on an array of tasks (compressibility, incompressibility, prompt image alignment, and LAION aesthetic score) proposed by DDPO, \alg{} outperforms DDPO in tested compression, incompression, and aesthetic tasks in training time, inference time, and sample complexity (i.e., total reward of the learned policy versus number of reward model queries used in training) (\cref{sec:experiments}). \looseness=-1

Our contributions in this work are as follows:
\begin{itemize}
    \vspace{-10px}
    \item In our experiments, we find that \alg{} has \textbf{\textit{faster training}} and \textbf{\textit{faster inference}} than existing methods.
    \item Further, that \alg{}, in our experiments, enjoys \textbf{\textit{better performance}} on most tasks under the tested reward models than existing methods.
\end{itemize}

\section{Related Works}
\label{sec:related_works}

\paragraph{Diffusion Models}
Diffusion models are a popular family of image generative models which progressively map noise to data \citep{sohl2015deep}. Such models generate high quality images \citep{ramesh2021zero, saharia2022photorealistic} and videos \citep{ho2022imagen, singer2022make}. Recent work with diffusion models has also shown promising directions in harnessing their power for other types of data such as robot trajectories and 3d shapes \citep{janner2022planning, zhou20213d}. However, the iterative inference procedure of progressively removing noise yields slow generation time.

\paragraph{Consistency Models}
Consistency models \citep{song2023consistency} are another family of generative models which directly map noise to data via the consistency function . Such a function provides faster inference generation as one can predict the image from randomly generated noise in a single step. Consistency models also offer a more fine-tuned trade-off between inference time and generation quality as one can run the multistep inference process (\cref{alg:mutlistep-cm}, in \cref{sec:append_cm}) which is described in detail in \cref{sec:inference-for-cm}. Prior works have also focused on training the consistency function in latent space \citep{luo2023latent} which allows for large, high quality text-to-image consistency model generations. Sometimes, such generations are not aligned with the downstream for which they will be used. The remainder of this work will focus on aligning consistency models to fit downstream preferences, given a reward function.

\paragraph{RL for Diffusion Models}
Popularized by \cite{black2024ddpo, fan2023DPOK}, training diffusion models with reinforcement learning requires treating the diffusion model inference sequence as a Markov decision process. Then, by treating the score function as the policy and updating it with a modified PPO algorithm \citep{schulman2017proximal}, one can learn a policy (which in this case is a diffusion model) that optimizes for a given downstream reward. Further work on RL fine-tuning has looked into entropy regularized control to avoid reward hacking and maintain high quality images \citep{uehara2024finetuning}. Another line of work uses deterministic policy gradient methods to directly optimize the reward function when the reward function is differentiable \citep{prabhudesai2023aligning}. Note that when reward function is differentiable, we can instantiate a deterministic policy gradient method in \alg{} as well. We focus on REINFORCE \citep{williams1992simple} style policy gradient methods for the purpose of optimizing a black-box, non-differentiable reward functions.

\section{Preliminaries}
\label{sec:preliminaries}

We provide some preliminary information on reinforcement learning, diffusion and consistency models, and discuss the application of reinforcement learning to diffusion models. Also note that we abuse notation and use $t$ to mean one of two things: the timestep along the diffusion trajectory or the timestep corresponding to the RL trajectory.

\subsection{Reinforcement Learning}
\label{subsec:rl_prelim}
We model our sequential decision process as a finite horizon Markov Decision Process (MDP), $\Mcal = (\Scal, \Acal, P, R, \mu, H)$. In this tuple, we define our state space $\Scal$, action space $\Acal$, transition function $P: \Scal \times \Acal \to \Delta(\Scal)$, reward function $R: \Scal \times \Acal \to \RR$, initial state distribution $\mu$, and horizon $H$. At each timestep $t$, the agent observes a state $s_t\in\Scal$, takes an action according to the policy $a_t\sim \pi(a_t | s_t)$ and transitions to the next state $s_{t+1}\sim P(s_{t+1} | s_t, a_t)$. After $H$ timesteps, the agent produces a trajectory as a sequence of states and actions $\tau = \left(s_0, a_0, s_1, a_1, \ldots, s_H, a_H\right)$. Our objective is to learn a policy $\pi$ that maximizes the expected cumulative reward over trajectories sampled from $\pi$, 
\[
    \Jcal_{RL} (\pi) = \EE_{\tau \sim p(\cdot | \pi)}  \left[\sum_{t=0}^H  R(s_t,a_t) \right].
\]

\subsection{Diffusion and Consistency Models}
\label{subsec:models_prelim}
Generative models are designed to match a model with the data distribution, such that we can synthesize new data points at will by sampling from the distribution. 
Diffusion models belong to a novel type of generative model that characterizes the probability distribution using a score function rather than a density function. 
Specifically, it produces data by gradually modifying the data distribution and subsequently generating samples from noise through a sequential denoising step.
More formally, we start with a distribution of data $p_\text{data}(\bx)$ and noise it according to the stochastic differential equation (SDE) \citep{song2020score}:
\[
 \text{d}\bx = \bm\mu (\bx_t, t) \text{d}t + \bm\sigma(t) \text{d}\bw
\]
for a given $t \in [0,T]$, fixed constant $ T > 0$, and with the drift coefficient $\bm\mu(\cdot, \cdot)$, diffusion coefficient $\bm\sigma(\cdot)$, and $\{\bw\}_{t\in [0,T]}$ being a Brownian motion. Letting $p_0(\bx) = p_\text{data}(\bx)$ and $p_t(x)$ be the marginal distribution at time $t$ induced by the above SDE, as shown in \cite{song2020score}, there exists an ODE (also called a \textit{probability flow}) whose induced distribution at time $t$ is also $p_t(\bx)$. In particular:
\[
\text{d}x_t = \left[ \bm\mu(\bx_t, t) - \frac12 \bm\sigma(t)^2 \nabla \log p_t (\bx_t) \right] \text{d}t
.\]

The term $\nabla \log p_t (\bx_t)$ is also known as the \textit{score function} \citep{song2019generative, song2020score}. When training a diffusion models in such a setting, one uses a technique called \textit{score matching} \citep{dinh2016density, vincent2011connection} in which one trains a network to approximate the score function and then samples a trajectory with an ODE solver. Once we learn such a neural network that approximates the score function, we can generate images by integrating the above ODE backward in time from $T$ to $0$, with $\bx_T \sim p_T$ which is typically a tractable distribution (e.g., Gaussian in most diffusion model formulations).

This technique is clearly bottle-necked by the fact that during generation, one must run a ODE solver backward in time (from $T$ to $0$) for a large number of steps in order to obtain competitive samples \citep{song2023consistency}. To alleviate this issue, \cite{song2023consistency} proposed \textit{consistency models} which aim to directly map noisy samples to data. The goal becomes instead to learn a \textit{consistency function} on a given probability flow. The aim of this function is that for any two $t,t' \in [\epsilon, T]$, the two samples along the probability flow ODE, they are mapped to the same image by the consistency function: $f_\theta(\bx_t, t) = f_\theta(\bx_{t'}, t') = \bx_\epsilon$ where $\bx_\epsilon$ is the solution of the ODE at time $\epsilon$. At a high level, this consistency function is trained by taking two adjacent timesteps and minimizing the consistency loss $d(f_\theta(\bx_t, t), f_\theta(\bx_{t'}, t'))$, under some image distance metric $d(\cdot, \cdot)$. To avoid the trivial solution of a constant, we also set the initial condition to $f_\theta(\bx_\epsilon, \epsilon) = \bx_\epsilon$.

\paragraph{Inference in consistency models} \label{sec:inference-for-cm} After a model is trained, one can then trade inference time for generation quality with the multi-step inference process given in \cref{sec:append_cm}, \cref{alg:mutlistep-cm}. At a high level, the multistep consistency sampling algorithm first partitions the probability flow into $H+1$ points ($T = \tau_0 > \tau_1 > \tau_2 \ldots > \tau_H = \epsilon$). Given a sample $x_T\sim p_T$, it then applies the consistency function $f_\theta$ at $(x_T,T)$ yielding $\wh \bx_0$. To further improve the quality of $\wh \bx_0$, one can add noise ($\bx \sim \Ncal(0,1)$) back to $\wh \bx_0$ using the equation
$\wh \bx_{\tau_n} \gets \wh\bx_0  +\sqrt{\tau_n^2 - \tau_H^2} \bz$, and then again apply the consistency function to $(\wh \bx_{\tau_n},\tau_n)$, getting $\wh \bx_0$. One can repeat this process for a few more steps until the quality of the generation is satisfied. %
For the remainder of this work, we will be referring to sampling with the multi-step procedure. We also provide more details when we introduce \alg{} later.

\subsection{Reinforcement Learning for Diffusion Models}
\label{subsec:rl_for_diffusion}

\citet{black2024ddpo} and \citet{fan2023DPOK} formulated the training and fine-tuning of conditional diffusion probabilistic models \citep{sohl2015deep, ho2020denoising} as an MDP. \citet{black2024ddpo} defined a class of algorithms, Denoising Diffusion Policy Optimization (DDPO), that optimizes for arbitrary reward functions to improve guided fine-tuning of diffusion models with RL. 
\paragraph{Diffusion Model Denoising as MDP} Conditional diffusion probabilistic models condition on a context $\bc$ (in the case of text-to-image generation, a prompt).
As introduced for DDPO, we map the iterative denoising procedure to the following MDP $\Mcal = (\Scal, \Acal, P, R, \mu, H)$. Let $r(\bs,\bc)$ be the task reward function. Also, note that the probability flow proceeds from $x_T \to x_0$. Let $T = \tau_0 > \tau_1 > \tau_2 \ldots > \tau_H = \epsilon$ be a partition of the probability flow into intervals:

\begin{align*}
    \bs_t &\defeq (\bc, \tau_t, x_{\tau_t}) &&\pi(\ba_t | \bs_t) \defeq p_{\theta}\left(x_{\tau_{t+1}} | x_{\tau_t}, \bc\right) &&P(\bs_{t+1} | \bs_t, \ba_t) \defeq (\delta_{\bc}, \delta_{\tau_{t+1}}, \delta_{x_{\tau_{t+1}}})\\
    \ba_t &\defeq x_{\tau_{t+1}} &&\mu \defeq \left(p(\bc), \delta_{\tau_0}, \Ncal(0, I)\right) &&R(\bs_t, \ba_t) = \begin{cases}r(\bs_t, \bc) & \text{if } t=H \\ 0 & \text{otherwise}\end{cases}
\end{align*} 

where $\delta_y$ is the Dirac delta distribution with non-zero density at $y$. In other words, we are mapping images to be states, and the prediction of the next state in the denosing flow to be actions. Further, we can think of the deterministic dynamics as letting the next state be the action selected by the policy. Finally, we can think of the reward for each state being $0$ until the end of the trajectory when we then evaluate the final image under the task reward function. 

This formulation permits the following loss term:
\[
\Lcal_{\text{DDPO}} = \EE_\Dcal \sum_{t=1}^T \left[ \min\left\{ r(\bx_0, \bc) \frac{p_\theta (\bx_{t-1} | \bx_t, \bc)} {p_{\theta_\text{old}} (\bx_{t-1} | \bx_t, \bc)} , r(\bx_0, \bc) \texttt{clip} \left( \frac{p_\theta (\bx_{t-1} | \bx_t, \bc)} {p_{\theta_\text{old}} (\bx_{t-1} | \bx_t, \bc)}, 1-\eps, 1+\eps \right)   \right\} \right]
\]
where clipping is used to ensure that when we optimize $p_{\theta}$, the new policy stay close to $p_{\theta_{old}}$, a trick popularized by the well known algorithm Proximal Policy Optimization (PPO) \citep{schulman2017proximal}. However, one could easily replace this with other policy gradient optimizers like \cite{gao2024rebel}.

In diffusion models (and in our experiments for DDPO), horizon $H$ is usually set as 50 or greater and time $T$ is set as $1000$. A small step size is chosen for the ODE solver to minimize error, ensuring the generation of high-quality images as demonstrated by \cite{ho2020denoising}. Due to the long horizon and sparse rewards, training diffusion models using reinforcement learning can be challenging.

\section{Reinforcement Learning for Consistency Models}
\label{sec:methods}

To remedy the long inference horizon that occurs during the MDP formulation of diffusion models, we instead frame consistency models as an MDP. We let $H$ also represent the horizon of this MDP. Just as we do for DDPO, we partition the entire probability flow ($[0,T]$) into segments, $T = \tau_0 > \tau_1 > \ldots > \tau_H = \epsilon$. In this section, we denote $t$ as the discrete time step in the MDP, i.e., $t \in \{0,1,\dots, H\}$, and $\tau_t$ is the corresponding time in the continuous time interval $[0,T]$. We now present the consistency model MDP formulation. 

\begin{figure}
    \centering
    \includegraphics[width=0.7\textwidth]{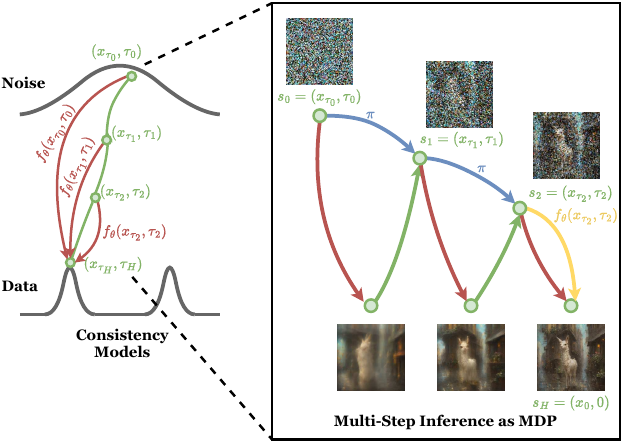}
    \caption{\textbf{Consistency Model As MDP: }In this instance, $H=3$. Here we first start at a randomly sampled noised state $s_0 \sim \left(\Ncal(0, I), \delta_{\tau_0}, p(\bc)\right)$. We then follow the policy by first plugging the state into the the consistency model (\textcolor{red}{red line}) and then noising the image back to $\tau_1$ (\textcolor{dark-green}{green line}). This gives us $a_0$ which, based off of the transition dynamics becomes $s_1$ (\textcolor{dark-green}{green circle}). We then transition from $s_1$ by applying $\pi(\cdot)$, which applies the consistency function to $\wh x_0$ and then noises up to $\tau_2$. To calculate the end of trajectory reward, we apply the consistency function one more time (\textcolor{yellow}{yellow line}) to get a final approximation of $\wh x_0$ and apply the given reward function to this image. Note that the red and green lines on both sides of the diagram represent the same thing.}
    \label{fig:rlcm-rollout-diagram}
\end{figure}

\paragraph{Consistency Model Inference as MDP} We reformulate the multi-step inference process in a consistency model (\cref{alg:mutlistep-cm}) as an MDP: %
\begin{align*}
    \bs_t &\defeq (\bx_{\tau_t}, \tau_t, \bc) &&\pi(\ba_t | \bs_t) \defeq f_{\theta}\left(\bx_{\tau_t}, \tau_t, \bc \right) + Z &&P(\bs_{t+1} | \bs_t, \ba_t) \defeq (\delta_{\bx_{\tau_{t+1}}}, \delta_{\tau_{t+1}}, \delta_{\bc})\\
    \ba_t &\defeq \bx_{\tau_{t+1}} &&\mu \defeq \left(\Ncal(0, I), \delta_{\tau_0}, p(\bc)\right) &&R_H(\bs_H) = r(f_\theta(\bx_{\tau_H}, \tau_H, \bc), \bc)
\end{align*}
where is $Z = \sqrt{\tau_t^2 - \tau_H^2}\bz$ which is noise from \cref{alg:noise-eqn} of \cref{alg:mutlistep-cm}. Further, where $r(\cdot, \cdot)$ is the reward function that we are using to align the model and $R_H$ is the reward at timestep $H$. At other timesteps, we let the reward be $0$. We can visualize this conversion from the multistep inference in \cref{fig:rlcm-rollout-diagram}.

Modeling the MDP such that the policy $\pi(s) \defeq f_\theta(\bx_{\tau_t}, \tau_t, \bc) +Z$ instead of defining $\pi(\cdot)$ to be the consistency function itself has a major benefit in the fact that this gives us a stochastic policy instead of a deterministic one. This allows us to use a form of clipped importance sampling like \cite{black2024ddpo} instead of a deterministic algorithm (e.g. DPG \citep{silver2014deterministic}) which we found to be unstable and in general is not unbiased. Thus a policy is made up of two parts: the consistency function and noising with Gaussian noises. The consistency function takes the form of the red arrows in \cref{fig:rlcm-rollout-diagram} whereas the noise is the green arrows. In other words, our policy is a Gaussian policy whose mean is modeled by the consistency function $f_\theta$, and covariance being $(\tau_t^2 - \epsilon^2) \mathbf{I}$ (here $\mathbf{I}$ is an identity matrix).  Notice that in accordance with the sampling procedure in \cref{alg:mutlistep-cm}, we only noise part of the trajectory. Note that the final step of the trajectory is slightly different. In particular, to calculate the final reward, we just apply the consistency function (red/yellow arrrow) and obtain the final reward.

\paragraph{Policy Gradient \alg{}} We can then instantiate RLCM with a policy gradient optimizer, in the spirit of \cite{black2024ddpo, fan2023DPOK}. Our algorithm is described in \cref{alg:pg-rlcm}.
\begin{algorithm}[tb!]
\caption{Policy Gradient Version of \alg{}}\label{alg:pg-rlcm}
\begin{algorithmic}[1]
\State \textbf{Input:} Consistency model policy $\pi_\theta = f_\theta(\cdot,\cdot) + Z$, finetune horizon $H$, prompt set $\Pcal$, batch size $b$, inference pipeline $P$
\For{$i = 1$ \textbf{to} $M$}
\State Sample $b$ contexts from $\Ccal$, $\bc \sim \Ccal$.
\State $\bm{x}_0 \gets P(f_\theta, H, \bc)$ \Comment{where $\bm{x}_0$ is the batch of images }
\State Normalize rewards $r(\bx_0,\bc)$ per context
\State Split $\bm{x}_0$ into $k$ minibatches.
\For{minibatch $m=0$ to $\texttt{ceil}(\texttt{length}(\bm{x}_0) / \texttt{minibatch\_size})$ }
\For{$t=0$ to $H$}
\State Update $\theta$ using rule: 
\[
    \nabla_\theta \left[ \min\left\{ r({\bx_{0,m}},\bc) \cdot \frac{\pi_{\theta_{m+1}} (a_t| s_t)} {\pi_{\theta_{m}} (a_t | s_t)} , r({\bx_{0,m}},\bc) \cdot \texttt{clip} \left( \frac{\pi_{\theta_{m+1}} (a_t | s_t)} {\pi_{\theta_{m}} (a_t | s_t)}, 1-\eps, 1+\eps \right)   \right\} \right]
\]
\EndFor
\EndFor
\EndFor
\State Output trained consistency model $f_\theta(\cdot, \cdot)$
\end{algorithmic}
\end{algorithm}
In practice we normalize the reward per prompt. That is, we create a running mean and standard deviation for each prompt and use that as the normalizer instead of calculating this per batch. This is because under certain reward models, the average score by prompt can vary drastically. 

\section{Experiments}
\label{sec:experiments}

In this section, we hope to investigate the performance and speed improvements of training consistency models rather than diffusion models with reinforcement learning. We compare our method to \ddpo{} \citep{black2024ddpo}, a state-of-the-art policy gradient method for finetuning diffusion models. First, we test how well \alg{} is able to both efficiently optimize the reward score and maintain the qualitative integrity of the pretrained generative model. We show both learning curves and representative qualitative examples of the generated images on tasks defined by \citet{black2024ddpo}. Next we show the speed and compute needs for both train and test time of each finetuned model to test whether \alg{} is able to maintain a consistency model's benefit of having a faster inference time. We then conduct an ablation study, incrementally decreasing the inference horizon to study \alg{}'s tradeoff for faster train/test time and reward score maximization. Finally, we qualitatively evaluate \alg{}'s ability to generalize to text prompts and subjects not seen at test time to showcase that the RL finetuning procedure did not destroy the base pretrained model's capabilities.

For fair comparison, both \ddpo{} and \alg{} finetune the Dreamshaper v7\footnote{\url{https://huggingface.co/Lykon/dreamshaper-7}} and its latent consistency model counterpart respectively\footnote{\url{https://huggingface.co/SimianLuo/LCM_Dreamshaper_v7}} \citep{luo2023latent}. Dreamshaper v7 is a finetune of stable diffusion \citep{rombach2022stable}. For \ddpo{}, we used the same hyperparameters and source code\footnote{\url{https://github.com/kvablack/ddpo-pytorch}}\citep{black2024ddpo} provided by the authors. We found that the default parameters performed best when testing various hyperparamters. Please see \cref{sec:append_hyper} for more details on the parameters we tested. 

\paragraph{Compression} The goal of compression is to minimize the filesize of the image. Thus, the reward received is equal to the negative of the filesize when compressed and saved as a JPEG image. The highest rated images for this task are images of solid colors. The prompt space consisted of 398 animal categories.

\begin{figure}[hb!]
    \centering
    \includegraphics[width=\textwidth]{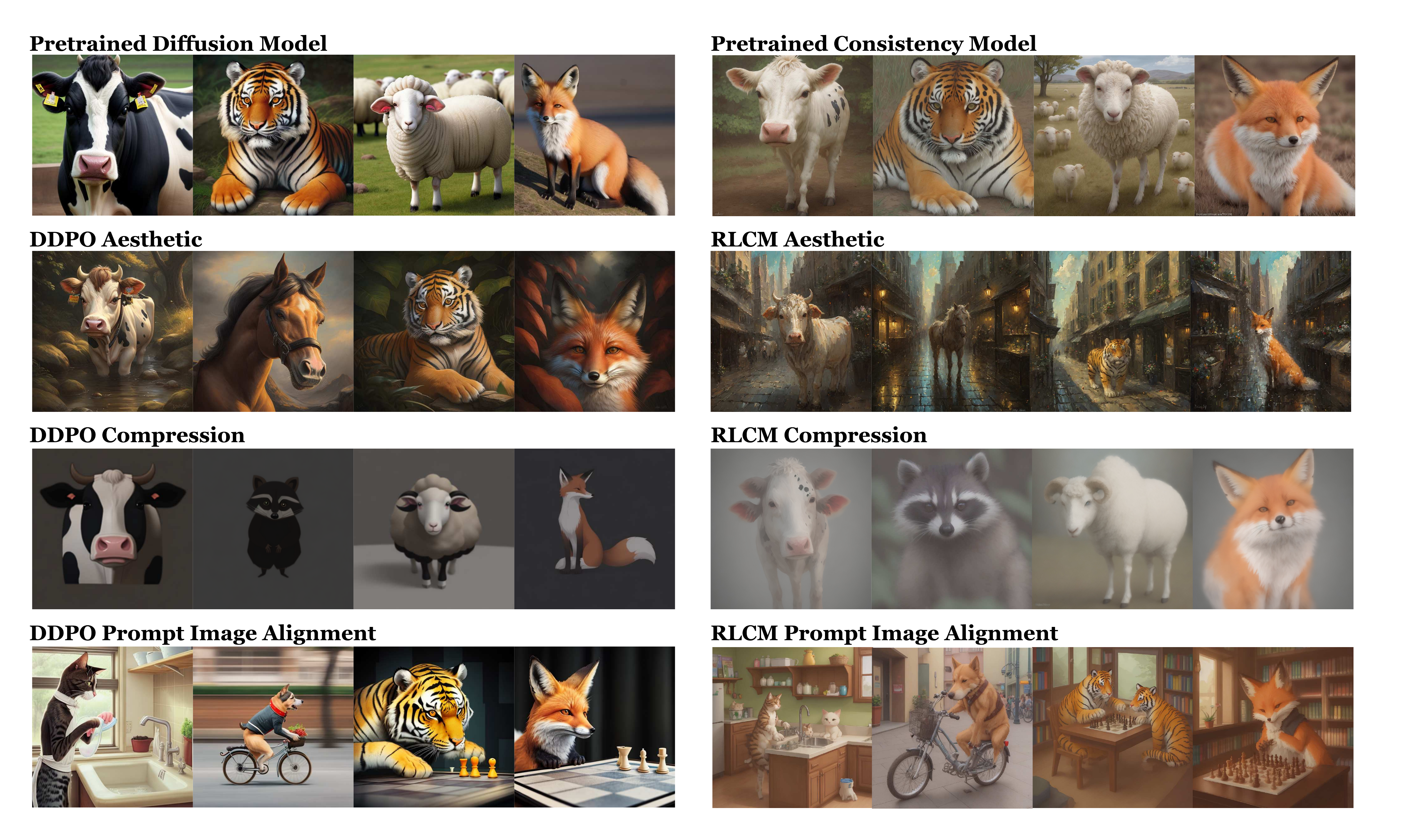}
    \caption{\textbf{Qualitative Generations: }Representative generations from the pretrained models, \ddpo{}, and \alg{}. Across all tasks, we see that \alg{} is able to finetune output of the model to fit specific reward functions. Due to the lack of regularization to the pretrained model, some artifacts (seen in the compression task) and significant similarity in output are indeed seen).}
    \label{fig:main-qual-result}
\end{figure}

\paragraph{Incompression} Incompression has the opposite goal of compression: to make the filesize as large as possible. The reward function here is just the filesize of the saved image. The highest rated mages for this task are random noise. Similar to the comparison task, this task's prompt space consisted of 398 animal categories.

\paragraph{Aesthetic} The aesthetic task is based off of the LAION Aesthetic predictor \citep{schumman2022} which was trained on 176,000 human labels of aesthetic quality of images. This aesthetic predictor is a MLP on top of CLIP embeddings \citep{radford2021learning}. The images which produce the highest reward are typically artwork. This task has a smaller set of 45 animals as prompts.\looseness=-1

\paragraph{Prompt Image Alignment}  We use the same task as \cite{black2024ddpo} in which the goal is to align the prompt and the image more closely without human intervention. This is done through a procedure of first querying a LLaVA model \citep{liu2023llava} to determine what is going  on in the image and taking that response and computing the BERT score \citep{zhang2019bertscore} similarity to determine how similar it is to the original prompt. This values is then used as the reward for the policy gradient algorithm.

\begin{figure}[t!]
    \centering
    \includegraphics[width=\textwidth]{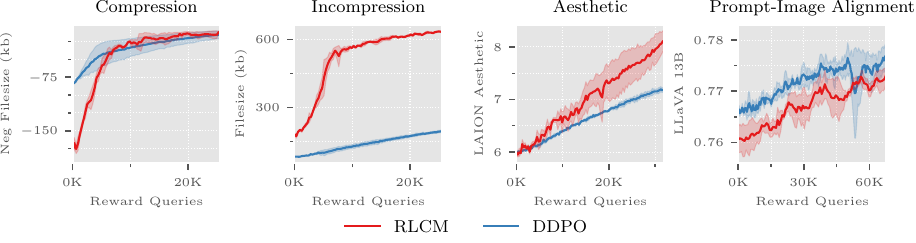}
    \caption{\textbf{Learning Curves: }Training curves for \alg{} and \ddpo{} by number of reward queries on compressibility, incompressibility, aesthetic, and prompt image alignment. We plot three random seeds for each algorithm and plot the mean and standard deviation across those seeds. \alg{} seems to produce either comparable or better reward optimization performance across these tasks.}
    \vspace{-4mm}
    \label{fig:main-plot}
\end{figure}

\begin{figure}[t!]
    \centering
    \includegraphics[width=\textwidth]{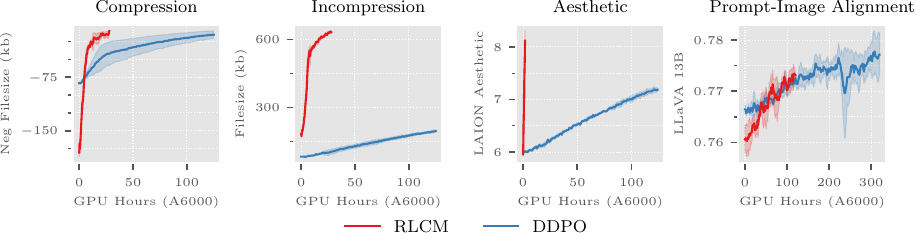}
    \caption{\textbf{Training Time: }Plots of performance by runtime measured by GPU hours. We report the runtime on four NVIDIA RTX A6000 across three random seeds and plot the mean and standard deviation. We observe that in all tasks \alg{} noticeably reduces the training time while achieving comparable or better reward score performance.}
    \label{fig:main-plot-time}
\end{figure}

\begin{figure}[t!]
    \centering
    \includegraphics[width=\textwidth]{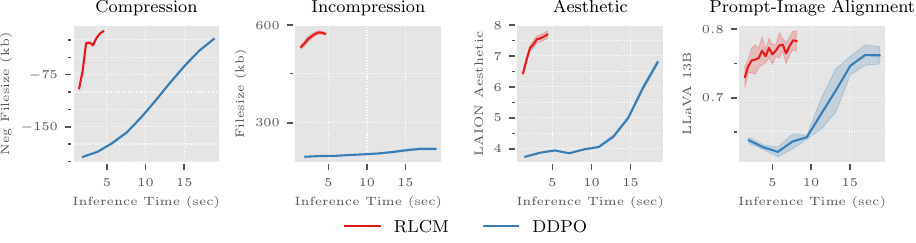}
    \caption{\textbf{Inference Time: }Plots showing the inference performance as a function of time taken to generate. For each task, we evaluated the final checkpoint obtained after training and measured the average score across 100 trajectories at a given time budget on 1 NVIDIA RTX A6000 GPU. We report the mean and std across three seeds for every run. Note that for \alg{}, we are able to achieve high scoring trajectories with a smaller inference time budget than \ddpo{}. Final reward values may differ from previous plots due to random selection of prompts used for measurement.}
    \label{fig:main-plot-time-test}
\end{figure}
\subsection{\alg{} vs. \ddpo{} Performance Comparisons}
Following the sample complexity evaluation proposed in \cite{black2024ddpo}, we first compare \ddpo{} and \alg{} by measuring how fast they can learn based on the number of reward model queries.  As shown in \cref{fig:main-plot}, \alg{} has better performance on three out of four of our tested tasks in terms of number of reward queries. Note that for the prompt-to-image alignment task, the initial consistency model finetuned by \alg{} has lower performance than the initial diffusion model trained by \ddpo{}. \alg{} is able to close the performance gap between the consistency and diffusion model through RL finetuning\footnote{It is possible that this performance difference on the compression and incompression tasks are due to the consistency models default image being larger. However, in the prompt image alignment and aesthetic tasks, we resized the images before reward calculation.}. \cref{fig:main-qual-result} demonstrates that similar to \ddpo{}, \alg{} is able to train its respective generative model to adapt to various styles just with a reward signal without any additional data curation or supervised finetuning.

\subsection{Train and Test Time Analysis}

To show faster training advantage of the proposed RLCM, we compare to DDPO in terms of training time in \cref{fig:main-plot-time}. Here we experimentally find that \alg{} has a significant advantage to DDPO in terms of the number of GPU hours required in order to achieve similar performance. On all tested tasks \alg{} reaches the same or greater performance than \ddpo{}, notably achieving a x17 speedup in training time on the Aesthetic task. This is most likely due to a combination of factors -- the shorter horizon in \alg{} leads to faster online data generation (rollouts in the RL training procedure) and policy optimization (e.g., less number of backpropagations for training the networks).

\cref{fig:main-plot-time-test} compares the inference time between \alg{} and \ddpo{}. For this experiment, we measured the average reward score obtained by a trajectory given a fixed time budget for inference. Similar to training, \alg{} is able to achieve a higher reward score with less time, demonstrating that \alg{} retains the computational benefits of consistency models compared to diffusion models. Note that a full rollout with \alg{} takes roughly a quarter of the time for a full rollout with \ddpo{}. 
\vspace{1mm}
\subsection{Ablation of Inference Horizon for \alg{}}
\begin{wrapfigure}[17]{l}{0.6\textwidth}
    \vspace{-4mm}
    \centering
    \includegraphics[width=0.58\textwidth]{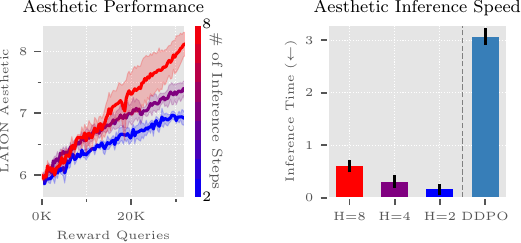}
    \caption{\textbf{Inference time vs Generation Quality: }We measure the performance of the policy gradient instantiation of \alg{} on the aesthetic task at 3 different values for the number of inference steps (left) in addition to measuring the inference speed in seconds with varied horizons (right). We report the mean and std across three seeds.}
    \label{fig:ablation}
\end{wrapfigure}

We further explore the effect of finetuning a consistency model with different inference horizons. That is we aimed to test \alg{}'s sensitivity to $H$.
As shown in \cref{fig:ablation} (left), increasing the number of inference steps leads to a greater possible gain in the reward. However, \cref{fig:ablation} (right) shows that this reward gain comes at the cost of slower inference time. This highlights the \textit{inference time vs generation quality} tradeoff that becomes available by using \alg{}. Nevertheless, regardless of the number of inference steps chosen, \alg{} enjoys faster inference time than diffusion model based baselines. 

\begin{figure}[b!]
    \centering
    \includegraphics[width=\textwidth]{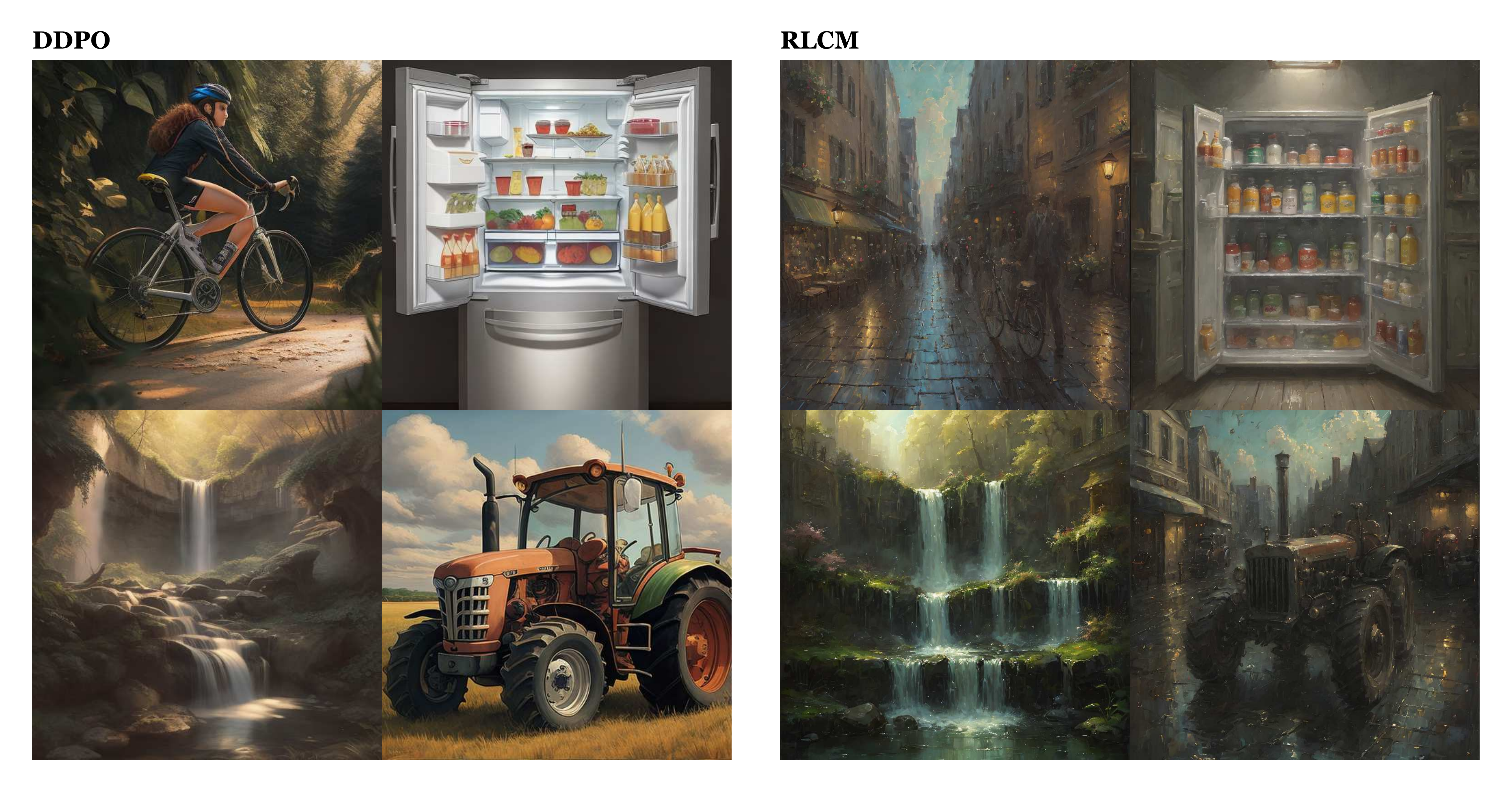}
    \caption{\textbf{Prompt Generalization: }We observe that \alg{} is able to generalize to other prompts without substantial decrease in aesthetic quality. The prompts used to test generalization are ``bike'', ``fridge'', ``waterfall'', and ``tractor''.}
    \label{fig:generalization}
    \vspace{-4mm}
\end{figure}
\subsection{Qualitative Effects on Generalization}
\vspace{-1.2mm}
We now test our trained models on new text prompts that do not appear in the training set. Specifically, we evaluated our trained models on the aesthetic task. As seen in \cref{fig:generalization} which consists of images of prompts that are not in the training dataset, the RL finetuning does not influence the ability of the model to generalize. 
We see this through testing a series of prompts (``bike'', ``fridge'', ``waterfall'', and ``tractor'') unseen during training.

\subsection{Convergence Results of Tasks}

To compare fairly to \cite{black2024ddpo}, we only train for only the same number of reward queries which means that in two tasks (Aesthetic and Prompt Image Alignment) convergence of the tasks is not shown. 

We trained DDPO and \alg{} for longer on the aesthetic task and observed that \alg{} asymptotically arrived at the approximate maximum reward value (value 10 is the maximum reward available in the training dataset for the reward model). For DDPO, when it runs longer (after 72 hours), it reaches a reward around 9.5, but unfortunately crashes.

We also attempted to run the text-image alignment task longer for DDPO, unfortunately we observed the same crashing behavior. We suspect that it is due to the fixed learning rate schedule used in the original DDPO codebase (note that for fair comparison, we use the original DDPO codebase with the default hyperparameters proposed by the authors of DDPO). Applying strategies like learning rate decay may stabilize DDPO, but it would require additional hyperparameter tuning for DDPO.

\subsection{Known Limitations}

The main known limitation observed throughout the use of \alg{} is overfitting to the reward function. Indeed, as seen in \cref{fig:main-qual-result}, unrealistic generations as seen in the compression task or extremely similar backgrounds like in the aesthetic task do arise. In cases where such overfitting is undesirable, a KL regularized loss which incorporates some measure of divergence between the currently trained model and the initial model will improve generations. However, this was not a focus of this work. 

\section{Conclusion and Future Directions}
\label{sec:conclusion}
\vspace{-1.2mm}

We present \alg{}, a fast and efficient RL framework to directly optimize a variety of rewards to train consistency models. We empirically show that \alg{} achieves better performance than a diffusion model RL baseline, \ddpo{}, on most tasks while enjoying the fast train and inference time benefits of consistency models. Finally, we provide qualitative results of the finetuned models and test their downstream generalization capabilities. 

There remain a few directions unexplored which we leave to future work. In particular, the specific policy gradient method presented uses a sparse reward. It may be possible to use a dense reward using the property that a consistency model always predicts to $x_0$. Another future direction is the possibility of creating a loss that further reinforces the consistency property, further improving the inference time capabilities of \alg{} policies. 

\section{Social Impact}
\label{sec:impact}
\vspace{-1.2mm}
We believe that it is important to urge caution when using such fine-tuning methods. In particular, these methods can be easily misused by designing a malicious reward function. We therefore urge this technology be used for good and with utmost care.

\section*{Code References}
\vspace{-1.2mm}
We use the following open source libraries for this work: NumPy \citep{harris2020array}, diffusers \citep{von-platen-etal-2022-diffusers}, and PyTorch \citep{paszke2017automatic}

\section*{Acknowledgements}
\label{sec:ack}
We would like to acknowledge Yijia Dai and Dhruv Sreenivas for their helpful technical conversations.

\bibliography{main}

\begin{thebibliography}{33}
\providecommand{\natexlab}[1]{#1}
\providecommand{\url}[1]{\texttt{#1}}
\expandafter\ifx\csname urlstyle\endcsname\relax
  \providecommand{\doi}[1]{doi: #1}\else
  \providecommand{\doi}{doi: \begingroup \urlstyle{rm}\Url}\fi

\bibitem[Agarwal et~al.(2021)Agarwal, Schwarzer, Castro, Courville, and Bellemare]{agarwal2021deep}
Rishabh Agarwal, Max Schwarzer, Pablo~Samuel Castro, Aaron~C Courville, and Marc Bellemare.
\newblock Deep reinforcement learning at the edge of the statistical precipice.
\newblock \emph{Advances in neural information processing systems}, 34:\penalty0 29304--29320, 2021.

\bibitem[Black et~al.(2024)Black, Janner, Du, Kostrikov, and Levine]{black2024ddpo}
Kevin Black, Michael Janner, Yilun Du, Ilya Kostrikov, and Sergey Levine.
\newblock Training diffusion models with reinforcement learning, 2024.

\bibitem[Deng et~al.(2009)Deng, Dong, Socher, Li, Li, and Fei-Fei]{deng2009imagenet}
Jia Deng, Wei Dong, Richard Socher, Li-Jia Li, Kai Li, and Li~Fei-Fei.
\newblock Imagenet: A large-scale hierarchical image database.
\newblock In \emph{2009 IEEE conference on computer vision and pattern recognition}, pp.\  248--255. Ieee, 2009.

\bibitem[Dinh et~al.(2016)Dinh, Sohl-Dickstein, and Bengio]{dinh2016density}
Laurent Dinh, Jascha Sohl-Dickstein, and Samy Bengio.
\newblock Density estimation using real nvp.
\newblock \emph{arXiv preprint arXiv:1605.08803}, 2016.

\bibitem[Fan et~al.(2023)Fan, Watkins, Du, Liu, Ryu, Boutilier, Abbeel, Ghavamzadeh, Lee, and Lee]{fan2023DPOK}
Ying Fan, Olivia Watkins, Yuqing Du, Hao Liu, Moonkyung Ryu, Craig Boutilier, Pieter Abbeel, Mohammad Ghavamzadeh, Kangwook Lee, and Kimin Lee.
\newblock Dpok: Reinforcement learning for fine-tuning text-to-image diffusion models.
\newblock \emph{arXiv preprint arXiv:2305.16381}, 2023.

\bibitem[Gao et~al.(2024)Gao, Chang, Zhan, Oertell, Swamy, Brantley, Joachims, Bagnell, Lee, and Sun]{gao2024rebel}
Zhaolin Gao, Jonathan~D Chang, Wenhao Zhan, Owen Oertell, Gokul Swamy, Kiant{\'e} Brantley, Thorsten Joachims, J~Andrew Bagnell, Jason~D Lee, and Wen Sun.
\newblock Rebel: Reinforcement learning via regressing relative rewards.
\newblock \emph{arXiv preprint arXiv:2404.16767}, 2024.

\bibitem[Harris et~al.(2020)Harris, Millman, van~der Walt, Gommers, Virtanen, Cournapeau, Wieser, Taylor, Berg, Smith, Kern, Picus, Hoyer, van Kerkwijk, Brett, Haldane, del R{\'{i}}o, Wiebe, Peterson, G{\'{e}}rard-Marchant, Sheppard, Reddy, Weckesser, Abbasi, Gohlke, and Oliphant]{harris2020array}
Charles~R. Harris, K.~Jarrod Millman, St{\'{e}}fan~J. van~der Walt, Ralf Gommers, Pauli Virtanen, David Cournapeau, Eric Wieser, Julian Taylor, Sebastian Berg, Nathaniel~J. Smith, Robert Kern, Matti Picus, Stephan Hoyer, Marten~H. van Kerkwijk, Matthew Brett, Allan Haldane, Jaime~Fern{\'{a}}ndez del R{\'{i}}o, Mark Wiebe, Pearu Peterson, Pierre G{\'{e}}rard-Marchant, Kevin Sheppard, Tyler Reddy, Warren Weckesser, Hameer Abbasi, Christoph Gohlke, and Travis~E. Oliphant.
\newblock Array programming with {NumPy}.
\newblock \emph{Nature}, 585\penalty0 (7825):\penalty0 357--362, September 2020.
\newblock \doi{10.1038/s41586-020-2649-2}.
\newblock URL \url{https://doi.org/10.1038/s41586-020-2649-2}.

\bibitem[Ho et~al.(2020)Ho, Jain, and Abbeel]{ho2020denoising}
Jonathan Ho, Ajay Jain, and Pieter Abbeel.
\newblock Denoising diffusion probabilistic models.
\newblock \emph{Advances in neural information processing systems}, 33:\penalty0 6840--6851, 2020.

\bibitem[Ho et~al.(2022)Ho, Chan, Saharia, Whang, Gao, Gritsenko, Kingma, Poole, Norouzi, Fleet, et~al.]{ho2022imagen}
Jonathan Ho, William Chan, Chitwan Saharia, Jay Whang, Ruiqi Gao, Alexey Gritsenko, Diederik~P Kingma, Ben Poole, Mohammad Norouzi, David~J Fleet, et~al.
\newblock Imagen video: High definition video generation with diffusion models.
\newblock \emph{arXiv preprint arXiv:2210.02303}, 2022.

\bibitem[Janner et~al.(2022)Janner, Du, Tenenbaum, and Levine]{janner2022planning}
Michael Janner, Yilun Du, Joshua~B Tenenbaum, and Sergey Levine.
\newblock Planning with diffusion for flexible behavior synthesis.
\newblock \emph{arXiv preprint arXiv:2205.09991}, 2022.

\bibitem[Liu et~al.(2023)Liu, Li, Wu, and Lee]{liu2023llava}
Haotian Liu, Chunyuan Li, Qingyang Wu, and Yong~Jae Lee.
\newblock Visual instruction tuning.
\newblock In \emph{NeurIPS}, 2023.

\bibitem[Luo et~al.(2023)Luo, Tan, Huang, Li, and Zhao]{luo2023latent}
Simian Luo, Yiqin Tan, Longbo Huang, Jian Li, and Hang Zhao.
\newblock Latent consistency models: Synthesizing high-resolution images with few-step inference, 2023.

\bibitem[Paszke et~al.(2017)Paszke, Gross, Chintala, Chanan, Yang, DeVito, Lin, Desmaison, Antiga, and Lerer]{paszke2017automatic}
Adam Paszke, Sam Gross, Soumith Chintala, Gregory Chanan, Edward Yang, Zachary DeVito, Zeming Lin, Alban Desmaison, Luca Antiga, and Adam Lerer.
\newblock Automatic differentiation in pytorch.
\newblock 2017.

\bibitem[Prabhudesai et~al.(2023)Prabhudesai, Goyal, Pathak, and Fragkiadaki]{prabhudesai2023aligning}
Mihir Prabhudesai, Anirudh Goyal, Deepak Pathak, and Katerina Fragkiadaki.
\newblock Aligning text-to-image diffusion models with reward backpropagation, 2023.

\bibitem[Radford et~al.(2021)Radford, Kim, Hallacy, Ramesh, Goh, Agarwal, Sastry, Askell, Mishkin, Clark, et~al.]{radford2021learning}
Alec Radford, Jong~Wook Kim, Chris Hallacy, Aditya Ramesh, Gabriel Goh, Sandhini Agarwal, Girish Sastry, Amanda Askell, Pamela Mishkin, Jack Clark, et~al.
\newblock Learning transferable visual models from natural language supervision.
\newblock In \emph{International conference on machine learning}, pp.\  8748--8763. PMLR, 2021.

\bibitem[Ramesh et~al.(2021)Ramesh, Pavlov, Goh, Gray, Voss, Radford, Chen, and Sutskever]{ramesh2021zero}
Aditya Ramesh, Mikhail Pavlov, Gabriel Goh, Scott Gray, Chelsea Voss, Alec Radford, Mark Chen, and Ilya Sutskever.
\newblock Zero-shot text-to-image generation.
\newblock In \emph{International Conference on Machine Learning}, pp.\  8821--8831. PMLR, 2021.

\bibitem[Rombach et~al.(2022)Rombach, Blattmann, Lorenz, Esser, and Ommer]{rombach2022stable}
Robin Rombach, Andreas Blattmann, Dominik Lorenz, Patrick Esser, and Bj\"orn Ommer.
\newblock High-resolution image synthesis with latent diffusion models.
\newblock In \emph{Proceedings of the IEEE/CVF Conference on Computer Vision and Pattern Recognition (CVPR)}, pp.\  10684--10695, June 2022.

\bibitem[Saharia et~al.(2022)Saharia, Chan, Saxena, Li, Whang, Denton, Ghasemipour, Gontijo~Lopes, Karagol~Ayan, Salimans, et~al.]{saharia2022photorealistic}
Chitwan Saharia, William Chan, Saurabh Saxena, Lala Li, Jay Whang, Emily~L Denton, Kamyar Ghasemipour, Raphael Gontijo~Lopes, Burcu Karagol~Ayan, Tim Salimans, et~al.
\newblock Photorealistic text-to-image diffusion models with deep language understanding.
\newblock \emph{Advances in Neural Information Processing Systems}, 35:\penalty0 36479--36494, 2022.

\bibitem[Schulman et~al.(2017)Schulman, Wolski, Dhariwal, Radford, and Klimov]{schulman2017proximal}
John Schulman, Filip Wolski, Prafulla Dhariwal, Alec Radford, and Oleg Klimov.
\newblock Proximal policy optimization algorithms.
\newblock \emph{arXiv preprint arXiv:1707.06347}, 2017.

\bibitem[Schumman(2022)]{schumman2022}
Chrisoph Schumman.
\newblock Laion aesthetics.
\newblock \url{https://laion.ai/blog/ laion- aesthetics/}, 2022.

\bibitem[Silver et~al.(2014)Silver, Lever, Heess, Degris, Wierstra, and Riedmiller]{silver2014deterministic}
David Silver, Guy Lever, Nicolas Heess, Thomas Degris, Daan Wierstra, and Martin Riedmiller.
\newblock Deterministic policy gradient algorithms.
\newblock In \emph{International conference on machine learning}, pp.\  387--395. Pmlr, 2014.

\bibitem[Singer et~al.(2022)Singer, Polyak, Hayes, Yin, An, Zhang, Hu, Yang, Ashual, Gafni, et~al.]{singer2022make}
Uriel Singer, Adam Polyak, Thomas Hayes, Xi~Yin, Jie An, Songyang Zhang, Qiyuan Hu, Harry Yang, Oron Ashual, Oran Gafni, et~al.
\newblock Make-a-video: Text-to-video generation without text-video data.
\newblock \emph{arXiv preprint arXiv:2209.14792}, 2022.

\bibitem[Sohl-Dickstein et~al.(2015)Sohl-Dickstein, Weiss, Maheswaranathan, and Ganguli]{sohl2015deep}
Jascha Sohl-Dickstein, Eric Weiss, Niru Maheswaranathan, and Surya Ganguli.
\newblock Deep unsupervised learning using nonequilibrium thermodynamics.
\newblock In \emph{International conference on machine learning}, pp.\  2256--2265. PMLR, 2015.

\bibitem[Song \& Ermon(2019)Song and Ermon]{song2019generative}
Yang Song and Stefano Ermon.
\newblock Generative modeling by estimating gradients of the data distribution.
\newblock \emph{Advances in neural information processing systems}, 32, 2019.

\bibitem[Song et~al.(2020)Song, Sohl-Dickstein, Kingma, Kumar, Ermon, and Poole]{song2020score}
Yang Song, Jascha Sohl-Dickstein, Diederik~P Kingma, Abhishek Kumar, Stefano Ermon, and Ben Poole.
\newblock Score-based generative modeling through stochastic differential equations.
\newblock \emph{arXiv preprint arXiv:2011.13456}, 2020.

\bibitem[Song et~al.(2023)Song, Dhariwal, Chen, and Sutskever]{song2023consistency}
Yang Song, Prafulla Dhariwal, Mark Chen, and Ilya Sutskever.
\newblock Consistency models.
\newblock \emph{arXiv preprint arXiv:2303.01469}, 2023.

\bibitem[Uehara et~al.(2024)Uehara, Zhao, Black, Hajiramezanali, Scalia, Diamant, Tseng, Biancalani, and Levine]{uehara2024finetuning}
Masatoshi Uehara, Yulai Zhao, Kevin Black, Ehsan Hajiramezanali, Gabriele Scalia, Nathaniel~Lee Diamant, Alex~M Tseng, Tommaso Biancalani, and Sergey Levine.
\newblock Fine-tuning of continuous-time diffusion models as entropy-regularized control, 2024.

\bibitem[Vincent(2011)]{vincent2011connection}
Pascal Vincent.
\newblock A connection between score matching and denoising autoencoders.
\newblock \emph{Neural computation}, 23\penalty0 (7):\penalty0 1661--1674, 2011.

\bibitem[von Platen et~al.(2022)von Platen, Patil, Lozhkov, Cuenca, Lambert, Rasul, Davaadorj, and Wolf]{von-platen-etal-2022-diffusers}
Patrick von Platen, Suraj Patil, Anton Lozhkov, Pedro Cuenca, Nathan Lambert, Kashif Rasul, Mishig Davaadorj, and Thomas Wolf.
\newblock Diffusers: State-of-the-art diffusion models.
\newblock \url{https://github.com/huggingface/diffusers}, 2022.

\bibitem[Williams(1992)]{williams1992simple}
Ronald~J. Williams.
\newblock Simple statistical gradient-following algorithms for connectionist reinforcement learning.
\newblock 8\penalty0 (3):\penalty0 229--256, 1992.
\newblock ISSN 1573-0565.
\newblock \doi{10.1007/BF00992696}.
\newblock URL \url{https://doi.org/10.1007/BF00992696}.

\bibitem[Xu et~al.(2022)Xu, Yu, Song, Shi, Ermon, and Tang]{xu2022geodiff}
Minkai Xu, Lantao Yu, Yang Song, Chence Shi, Stefano Ermon, and Jian Tang.
\newblock Geodiff: A geometric diffusion model for molecular conformation generation.
\newblock \emph{arXiv preprint arXiv:2203.02923}, 2022.

\bibitem[Zhang et~al.(2019)Zhang, Kishore, Wu, Weinberger, and Artzi]{zhang2019bertscore}
Tianyi Zhang, Varsha Kishore, Felix Wu, Kilian~Q Weinberger, and Yoav Artzi.
\newblock Bertscore: Evaluating text generation with bert.
\newblock \emph{arXiv preprint arXiv:1904.09675}, 2019.

\bibitem[Zhou et~al.(2021)Zhou, Du, and Wu]{zhou20213d}
Linqi Zhou, Yilun Du, and Jiajun Wu.
\newblock 3d shape generation and completion through point-voxel diffusion.
\newblock In \emph{Proceedings of the IEEE/CVF International Conference on Computer Vision}, pp.\  5826--5835, 2021.

\end{thebibliography}
\bibliographystyle{rlc}

\newpage
\appendix

\section{Consistency Models}
\label{sec:append_cm}
We reproduce the consistency model algorithm from \cite{song2023consistency}.
\begin{algorithm}[H]
\caption{Consistency Model Multi-step Sampling Procedure \citep{song2023consistency}}\label{alg:mutlistep-cm}
\begin{algorithmic}[1]
\State \textbf{Input:} Consistency model $\pi = f_\theta(\cdot,\cdot)$, sequence of time points $\tau_1 > \tau_2 > \ldots > \tau_{N-1}$, initial noise $\wh \bx_T$
\State $\bx \gets f(\wh \bx_T, T)$
\For{$n=1$ \textbf{to} N-1}
\State $\bz \sim \Ncal (\bf 0, \bf I)$
\State $\wh\bx_{\tau_n} \gets \bx + \sqrt{\tau_n^2 - \epsilon^2}\bz$ \label{alg:noise-eqn}
\State $x \gets f(\wh\bx_{\tau_n}, \tau_n)$
\EndFor 
\State \textbf{Output:} $\bx$
\end{algorithmic}
\end{algorithm}

\section{Experiment Details}
\subsection{Hyperparameters}
\begin{table*}[tbh!]
\centering
\resizebox{\textwidth}{!} {
\begin{tabular}{l|c|c|c|c}
\toprule
Parameters &Compression &Incompression &Aesthetic &Prompt Image Alignment \\\midrule
Advantage Clip Maximum &10 &10 &10 &10\\
Batches Per Epoch &10 &10 &10 &6\\
Clip Range &0.0001 &0.0001 &0.0001 &0.0001\\
Gradient Accumulation Steps &2 &2 &4 &20\\
Learning Rate &0.0001 &0.0001 &0.0001 &0.0001\\
Max Grad Norm &5 &5 &5 &5\\
Pretrained Model & Dreamshaper v7 &Dreamshaper v7 &Dreamshaper v7 &Dreamshaper v7\\
Number of Epochs &100 &100 &100 &118\\
Horizon (Number of inference steps) &8 &8 &8 &16\\
Number of Sample Inner Epochs &1 &1 &1 &5\\
Sample Batch Size (per GPU) &4 &4 &8 &8\\
Rolling Statistics Buffer Size &16 &16 &32 &32\\
Rolling Statistics Min Count &16 &16 &16 &16\\
Train Batch Size (per GPU) &2 &2 &2 &2\\
Number of GPUs & 4 & 4 & 4 & 3 \\
LoRA rank & 16 & 16 & 8 & 16 \\
LoRA $\alpha$ & 32 & 32 & 8 & 32 \\
Consistency Model Time Horizon & 1000 & 1000 & 1000 & 1000 \\
\bottomrule
\end{tabular}
}
\caption{Hyperparameters for all tasks (Compression, Incompression, Aesthetic, Prompt Image Alignment)}
\label{tab:param}
\end{table*}
We note that a 4th gpu was used for Prompt Image Alignment as a sever for the LLaVA \citep{liu2023llava} and BERT models \citep{zhang2019bertscore} to form the reward function.

\subsection{Hyperparameter Sweep Ranges}
\label{sec:append_hyper}
These hyperparameters were found via a sweep. In particular we swept the learning rate for values in the range [1e-5,3e-4]. Likewise we also swept the number of batches per epoch and gradient accumulation steps but found that increasing both of these values led to greater performance, at the cost of sample complexity. We also swept the hyperparameters for DDPO, our baseline, but found that the provided hyperparameters provided the best results. In particular we tried lower batch size to increase the sample complexity of DDPO but found that this made the algorithm unstable. Likewise, we found that increasing the number of inner epochs did not help performance. In fact, it had quite the opposite effect.

\subsection{Details on Task Prompts}
We followed \citep{black2024ddpo} in forming the prompts for each of the tasks. The prompts for incompression, compression, and aesthetic took the form of \texttt{[animal]}. For the prompt image alignment task, the prompt took the form of \texttt{a [animal] [task]} where the \texttt{a} was conjugated depending on the animal. The prompts for compression and incompression were the animal classes of Imagenet \citep{deng2009imagenet}. Aesthetic was a set of simple animals, and prompt image alignment used the animals from the aesthetic task and chose from the tasks: \texttt{riding a bike}, \texttt{washing the dishes}, \texttt{playing chess}.

\section{Statistical Testing on Results}
Following \cite{agarwal2021deep}, we compute 95\% stratified bootstrap confidence intervals of the IQM, Mean, Median, and Optimality gap over the 4 tasks tested. We find that there is a statistically significant difference in rewards favoring RLCM for the mean, median, and optimality gap. There is slight overlap in the confidence intervals for the IQM.

\begin{figure}[H]
    \centering
    \includegraphics[width=\textwidth]{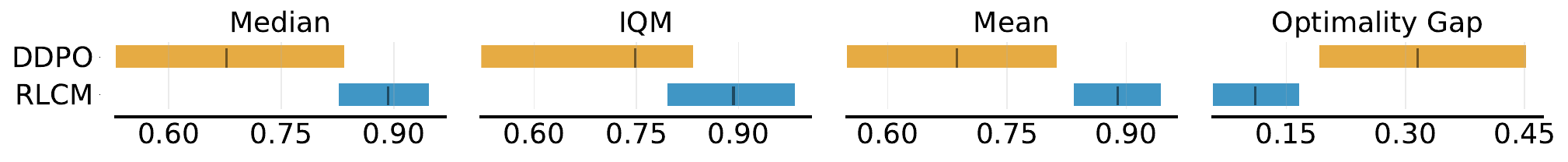}
    \caption{\textbf{Statistical Tests}: Stratified bootstrap confidence intervals and establish statistically significant difference in reward favoring RLCM.}
\end{figure}

\newpage
\section{Additional Samples from \alg{}}
We provide random samples from \alg{} at the end of training on aesthetic and prompt image alignment. Images from converged compression and incompression are relatively uninteresting and thus omitted.
\subsection{Aesthetic Task}
\begin{figure}[H]
    \centering
\includegraphics{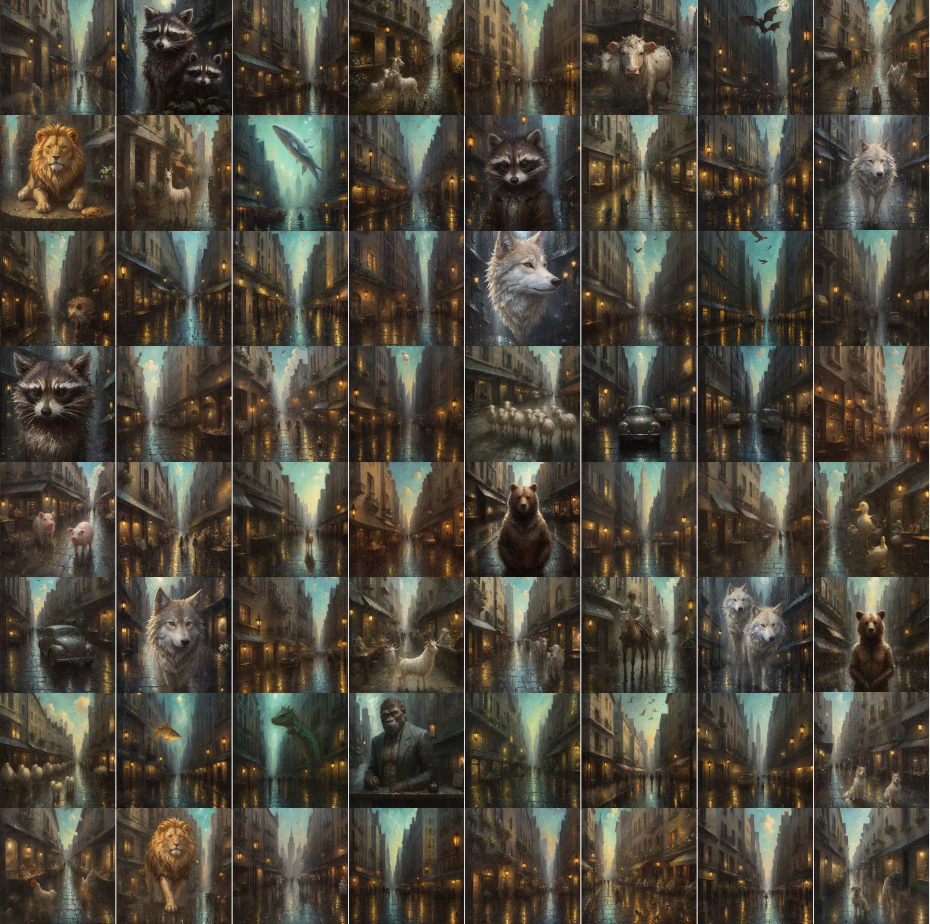}
\end{figure}
\newpage
\subsection{Prompt Image Alignment}
\begin{figure}[H]
    \centering
    \includegraphics{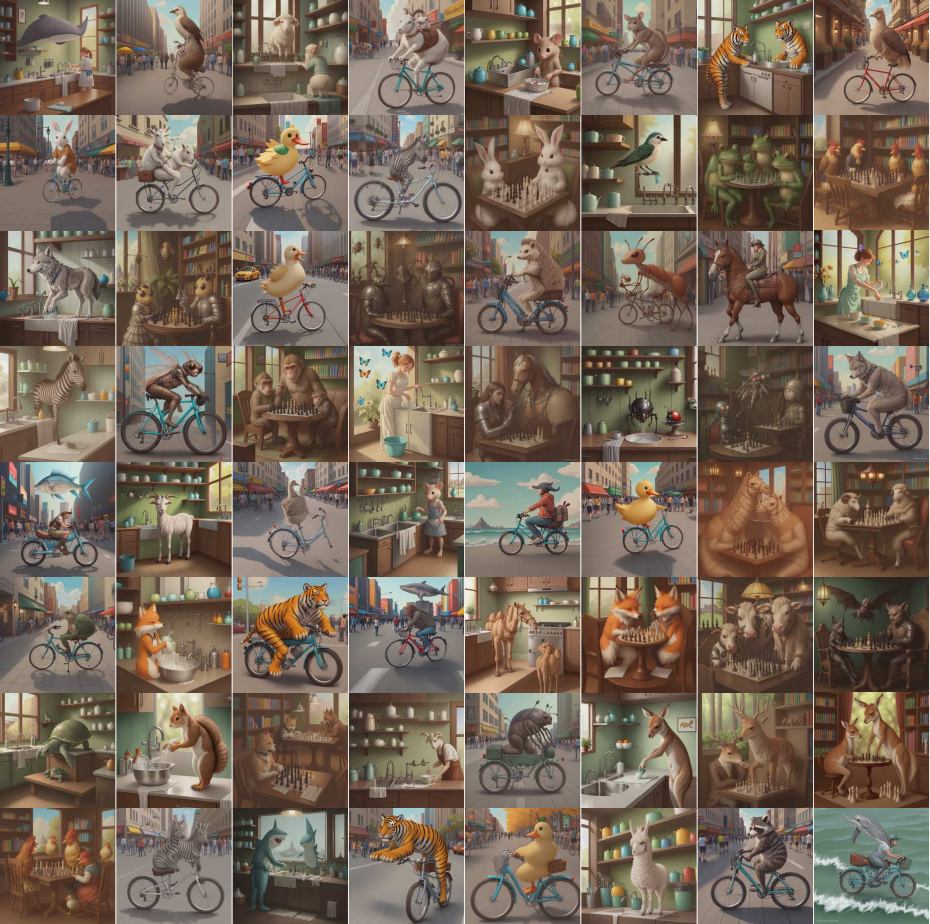}
\end{figure}
\end{document}